\documentclass[preprint,12pt]{elsarticle}

\usepackage{amssymb}
\usepackage{amsmath}
\usepackage{placeins}
\usepackage{makecell}
\usepackage{subcaption}
\usepackage{tabularx}
\usepackage{tcolorbox}

\journal{Engineering Applications of Artificial Intelligence (EAAI)}

\begin{document}

\begin{frontmatter}



\title{Generating Realistic, Protocol-Compliant Maritime Radio Dialogues using Self-Instruct and Low-Rank Adaptation} 

\author[label1]{Gürsel Akdeniz}
\author[label1]{Emin Cagatay Nakilcioglu}
\affiliation[label1]{organization={Fraunhofer Center for Maritime Logistics and Services (CML)},
                    city={Hamburg},
                    country={Germany}}


\begin{abstract}
Very high frequency (VHF) radio miscommunication remains a major safety risk in maritime operations, with human factors accounting for over 58\% of recorded incidents in Europe between 2014 and 2023. Despite decades of operational use, VHF radio communications are still prone to noise, interference, linguistic variability, and the absence of real-time transcription, making procedural errors both frequent and difficult to correct. Developing artificial intelligence-assisted systems to support real-time communication and decision-making requires a considerable amount of high-quality maritime data, yet operational, regulatory, and privacy constraints render such datasets scarce.

This study introduces a compliance-aware Self-Instruct methodology for generating realistic maritime radio dialogues that conform to the International Maritime Organization’s Standard Marine Communication Phrases (SMCP). Our approach integrates a 26-filter verification pipeline directly into the iterative generation loop to enforce entity information accuracy, hallucination detection, SMCP-compliance, logical consistency, and linguistic diversity. We employ Low-Rank Adaptation for parameter-efficient fine-tuning, reducing computational overhead during training and enabling efficient deployment of the resulting models on resource-constrained maritime systems.

To assess dataset quality, we introduce a four-metric evaluation framework combining automated and expert assessments: \textit{Format Accuracy}, \textit{Information Accuracy}, \textit{Uniqueness}, and \textit{Logical Coherence}. Experiments using publicly available vessel, coastal  and Automatic Identification System (AIS) datasets demonstrate that the approach produces synthetically diverse, procedurally compliant, and operationally realistic dialogues. Although downstream applications such as automatic speech recognition and natural language processing are reserved for future work, the released code, datasets, and verification tools provide a reproducible foundation for artificial intelligence-assisted maritime safety and other safety-critical domains.

\end{abstract}



\begin{keyword}
Maritime Radio Communication, Large Language Models, Domain Adaptation, In-domain Data Generation, Synthetic Data Generation, Self-Instruct, Low-Rank Adaptation, Parameter-Efficient Fine-Tuning,  Evaluation Framework


\end{keyword}

\end{frontmatter}


\section{Introduction}
\label{sec:introduction}
The maritime industry underpins global commerce, facilitating over 80\% of international trade with volumes reaching 12.292 million tons in 2023 and projected to grow at an average annual rate of 2.4\% between 2025 and 2029 \cite{rmt23, rmt24}. However, this growth is accompanied by persistent safety challenges. Between 2014 and 2023, European authorities recorded 26,595 marine incidents resulting in 7,604 injuries and 650 fatalities, with collisions and flooding as the predominant incident types \cite{emsa24}. Critically, human factors accounted for 58.4\% of these incidents as the primary cause, far outweighing system or equipment failures at 25.9\% \cite{emsa24}. This disparity underscores that human-centered interventions, supported by intelligent decision-making systems, are essential for maritime safety.

VHF Radio communication remains the primary medium for ship-to-ship and ship-to-shore coordination, yet miscommunication during procedural tasks has emerged as a leading safety threat. Despite decades of use, VHF radio is still vulnerable to multiple degradation factors, including adverse weather conditions, intrinsic radio noise and static, vessel-generated interference, and linguistic challenges, including accents, dialects, and multilingual speakers \cite{english,brcko2024}. Moreover, radio communications are often not transcribed or logged in real time, forcing operators to rely on memory during high-stress situations. These vulnerabilities make procedural communication one of the most failure-prone components of maritime operations.

AI-assisted radio communication systems offer a promising avenue to mitigate these risks. Such systems combine multiple complementary capabilities: (1) Automatic Speech Recognition (ASR) for real-time transcription of noisy VHF communications; (2) Procedural Guidance Systems that support operators in formulating SMCP-compliant communications by suggesting standardized phrases aligned with the current operational context; and (3) Communication Monitoring Assistance that alerts operators to potential procedural deviations, enabling them to self-correct before critical communications are sent. These capabilities are designed not to replace operator judgment, but to augment human expertise by reducing cognitive load, compensating for the high-stress conditions under which maritime personnel operate, and providing decision support for procedurally complex scenarios. Effective implementation and deployment of such human-in-the-loop systems requires substantial volumes of realistic maritime communication data that strictly adheres to the International Maritime Organization's (IMO) Standard Marine Communication Phrases (SMCP), a standardized vocabulary of approximately 900 phrases designed to transcend linguistic and cultural barriers while ensuring unambiguous communication during emergencies \cite{imo2001}.

Advancing AI-enabled maritime safety interventions requires high-quality, domain-specific maritime communication datasets for training and validation. However, maritime radio communication datasets are rarely public and extremely difficult to acquire due to operational sensitivity, regulatory constraints, and the episodic nature of maritime voyages \cite{MARTIUS2024100114, nakilcioglu, 10888216}. This data scarcity fundamentally limits the development of downstream maritime AI applications and explains why prior maritime AI research has relied on limited, self-curated datasets or simulation-based approaches.

This scarcity motivates exploring synthetic data generation. Although rule-based template approaches can produce procedurally compliant dialogues, they lack contextual fluency, semantic diversity, and scenario coverage necessary for realistic AI training. Conversely, Large Language Models (LLMs) offer scalable, contextually authentic synthetic dialogue generation, enabling the creation of diverse maritime scenarios that reflect operational challenges. LLMs have demonstrated notable versatility across specialized domains \cite{raffel2023exploringlimitstransferlearning, lewis2019bartdenoisingsequencetosequencepretraining, devlin2019bertpretrainingdeepbidirectional, openai2024gpt4technicalreport, ye2023comprehensivecapabilityanalysisgpt3, grattafiori2024llama3herdmodels, workshop2023bloom176bparameteropenaccessmultilingual, jiang2023mistral7b}. In the maritime context, LLMs can be adapted to generate large-scale synthetic radio dialogues for direct use or conversion to audio via text-to-speech (TTS) for ASR training.

However, generating useful synthetic maritime data presents two key challenges. The first is compliance: general-purpose LLMs often fail to adhere to the strict regulatory and procedural standards defined by SMCP. While SMCP's standardized 900-phrase vocabulary ensures that communications are unambiguous and universally understood, general-purpose LLMs have no intrinsic knowledge of these constraints and frequently generate plausible-sounding but non-compliant utterances. The second is realism: LLMs frequently generate text that, while superficially plausible, lacks the operational authenticity, contextual coherence, domain-specific conventions, and real-world entities (vessel names, geographical coordinates, incident types etc.) necessary for realistic maritime scenarios. Training downstream AI systems on non-compliant or unrealistic synthetic data risks producing systems that fail during real-world emergencies, creating false confidence in operators and potentially exacerbating rather than mitigating human-factors incidents.

Several studies have explored LLM adaptation in specialized domains. Afzal et al. examined LLM adaptation for text summarization in science, health, and government contexts \cite{afzal2024adaptevalevaluatinglargelanguage}, while Wan et al. applied retrieval-augmented generation (RAG) to the Chinese legal domain \cite{wan2024reformulatingdomainadaptationlarge}. More recently, Nguyen et al. trained LLaMA 3.1 models (8B and 70B) with maritime-specific data to develop domain-enriched maritime LLMs \cite{nguyen2025llamarineopensourcemaritimeindustryspecific}. However, while RAG-based approaches excel in integrating static external knowledge, they typically do not embed compliance verification directly into the learning process, since their architectures decouple retrieval from model optimization \cite{lewis2021retrievalaugmentedgenerationknowledgeintensivenlp}. For maritime radio communications, such decoupling risks generating radio chatters that are plausible in language but fail to meet strict SMCP compliance or reflect real operational scenarios.

Iterative self-improvement methods like Self-Instruct \cite{wang2023selfinstructaligninglanguagemodels} are more suitable for bootstrapping domain knowledge through iterative generation and filtering without extensive manual annotation. By embedding regulatory compliance verification into the adaptation loop, compliance can become an intrinsic learned property rather than a post-hoc filtering step. To our knowledge, this dual requirement (compliance as an embedded constraint within iterative fine-tuning and operationally authentic realism) has been largely unexplored in the maritime AI literature.

We propose a compliance-aware Self-Instruct methodology that embeds regulatory verification into the iterative fine-tuning process. Unlike standard Self-Instruct, which relies on generic heuristic filtering, our approach integrates a novel 26-filter verification pipeline directly into the generation loop. This pipeline operates across multiple verification stages (entity information accuracy, hallucination detection, SMCP compliance, logical consistency and uniqueness) ensuring that only compliant, operationally authentic dialogues contribute to model training. The compliance constraints are thereby learned as intrinsic properties of the adapted model, rather than applied as external post-hoc filters. We combine this with parameter-efficient LoRA fine-tuning to enable deployment on resource-constrained maritime systems, which often operate in environments with limited computational capacity. Finally, we introduce a four-metric evaluation framework combining automated assessment (\textit{Format Accuracy, Information Accuracy, Uniqueness}) and expert judgment (\textit{Logical Coherence}), enabling rigorous quantification of synthetic data quality in terms of regulatory adherence, information fidelity, diversity, and real-world contextual realism.

Our validation strategy incorporates three components: (1) Public data sources including public AIS datasets as well as GeoNames and GSHHG databases to verify information accuracy and entity realism; (2) Expert maritime linguistic evaluation by domain specialists to assess logical coherence, procedural authenticity, and compliance with operational norms; and (3) Automated SMCP compliance verification against the official IMO SMCP handbook. All datasets rely exclusively on publicly available maritime sources. The complete dataset composition and preprocessing pipeline are detailed in Section \ref{sec:datasets}.


Our contributions are summarized as follows:
\begin{itemize}
    \item Compliance-aware Self-Instruct methodology with embedded 26-filter verification, generalizable to other regulatory domains (legal, medical, aviation).
    \item A high-quality synthetic maritime dataset containing logically coherent, SMCP-compliant distress calls with real vessel identifiers (IMO \& MMSI) and geographical data from public sources.
    \item A four-metric evaluation framework assessing synthetic data quality in terms of regulatory adherence, information fidelity, diversity, and operational realism.
    \item Open-source code, synthetic datasets, comprehensive filter specifications, and evaluation results enabling reproducibility and downstream applications.
\end{itemize}

The remainder of this paper is organized as follows. Section \ref{sec:methodology} details the compliance-aware Self-Instruct procedure, the 26-filter verification pipeline, and LoRA fine-tuning strategy. Section \ref{sec:results} presents quantitative and qualitative results comparing the base LLM, fine-tuned model, and evaluation metrics across the synthetic dataset. Section \ref{sec:limitations} discusses the limitations of this work and outlines potential directions for future research. Finally, Section \ref{sec:conclusion} provides a general conclusion and discusses the broader implications of this work. Open-source codebase, datasets, and evaluation metrics can be found in the GitHub repository\footnote{Code and data available at: https://github.com/Akdenizg/maritime-chatter-generation.}.

\section{Methodology}
\label{sec:methodology}
In this section, we detail our complete methodology, beginning with the open-source datasets and the base LLM used in the synthetic data generation process. We then provide a detailed explanation of our adaptation of the Self-Instruct method \cite{wang2023selfinstructaligninglanguagemodels} to the domain of maritime distress calls and outline the subsequent LLM fine-tuning procedure. Finally, we introduce the evaluation framework that consists of four metrics designed to assess the \textit{Format Accuracy}, \textit{Information Accuracy}, \textit{Uniqueness}, and \textit{Logical Coherency} of the generated calls. These metrics are crucial for quantifying the effectiveness of our approach and are used to conduct a comparative performance analysis between the base and the fine-tuned models.

\subsection{Datasets}
\label{sec:datasets}
To support the generation of maritime radio communication that incorporates real vessel and geographical data, we utilize four public datasets.

\begin{itemize}
    \item \textbf{Global Self-consistent, Hierarchical, High-Resolution Geography(GSHHG)  Database \cite{GSHHG}:} This dataset provides shoreline data. The full-resolution L1 data is used in the scope of this work.
    \item \textbf{Geonames \cite{geonames}:} GeoNames is a comprehensive geographical database containing over 25 million entries, including names, latitudes, longitudes, feature codes and country codes of geographical entities.
    \item \textbf{Marine Cadastre \cite{boem_noaa_marinecadastre}:} This dataset contains AIS data from the US Coast Guard starting from 2009.
    \item \textbf{Danish Maritime Authority AIS Dataset\cite{dma_ais}:} This dataset contains AIS data from the vicinity of Denmark, starting from the year 2006.
\end{itemize}

Vessel information, including names, call signs, vessel types and MMSI numbers, was extracted by combining AIS data from Marine Cadastre (from 01.01.2023 to 30.06.2024) and the Danish Maritime Authority AIS Dataset (data from 26.07.2024 to 14.02.2025). To prevent the over-representation of certain vessel types, we removed 5000 pleasure craft and 5000 sailing vessels from the Marine Cadastre, and ensured vessel type consistency during the combination process. Entries without a vessel name were removed, special characters were cleaned, and vessel types classified as "UNKNOWN" or "Other" were mapped to "Motor Vessel." The final vessel dataset contains 58855 vessels. We extracted names, latitudes, longitudes, feature codes, and country codes from Geonames, while also removing parentheses and brackets.

\subsection{LLM used in the study}
\label{sec:model_used}
Llama 3.1 8B base model was selected for this study, as it provides a suitable balance between performance and model size \cite{grattafiori2024llama3herdmodels}. 

\subsection{Self Instruct for Maritime Distress Call Training Dataset Generation}
\label{sec:pipeline}
Wang et al. developed the Self-Instruct method to create diverse datasets for fine-tuning large language models \cite{wang2023selfinstructaligninglanguagemodels}. This approach uses an iterative bootstrapping process, where a model is prompted to generate synthetic instructions and corresponding input-output pairs from a small set of manually created seed instances of instructions and input-output pairs. The generated data is then filtered to remove low-quality or similar instances before being used to fine-tune the original model. This process is repeated until a sufficiently large and diverse fine-tuning dataset is created. The resulting model, finetuned with this self-generated data, shows a significant improvement in its ability to follow instructions.

\begin{figure}[htb]
    \centering
    \includegraphics[width = \textwidth]{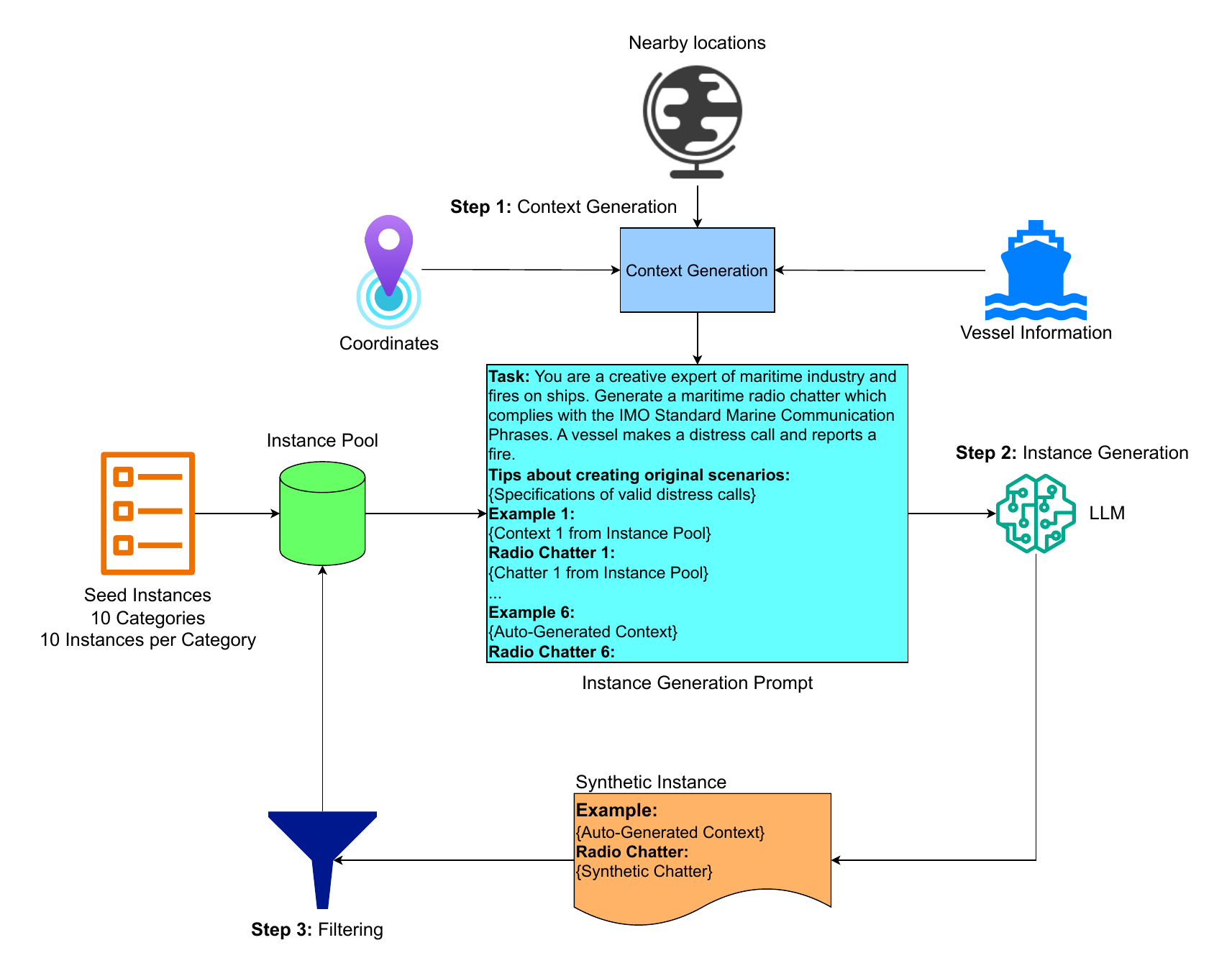}
    \caption{The Realistic Maritime Distress Call Generation Pipeline}
    \label{fig:pipeline}
\end{figure}

We adapted the Self-Instruct method to generate a training dataset for maritime distress calls. The overall process is illustrated in Figure \ref{fig:pipeline}. We began by manually creating a seed instances, $S$, consisting of 10 seed instances for each 10 distress categories from the SMCP. Each instance $S({I,X,C})$ comprises an instruction $I$, a context $X$, and a maritime distress call chatter $C$. The instructions correspond to the SMCP distress call categories, which are detailed in Table \ref{tab:instructions}. To ensure diversity, seed instances include 10 unique contexts and 10 unique chatters per category, for a total of 100 chatters. An example of a seed instance is provided in Table \ref{tab:example_seed}. The phrases that form the basis of these seed instances were derived from SMCP chapter A1. We intentionally omitted the category "Abandoning Vessel" as it contains a single phrase. However, to promote diversity, we also incorporated suitable phrases from various categories into other categories. 

Initially, we generate the context of the distress call from the datasets described in section \ref{sec:datasets}. Context includes information such as the coordinates of the vessel, nearby locations, distance to the nearest land, nearest landmark and vessel name. The layout of this context is explained in \ref{subsec:context generation}.

Subsequently, the synthetic context is appended to an instance generation prompt with 5 random examples from the instance pool $S$ to generate a synthetic instance. Table \ref{tab:instance_generation_template} outlines the template of the instruction for generating instances. Specific prompts used for each category can be found in the GitHub repository. Instance generation prompt with example instances is shown in table \ref{tab:instance_generation_prompt}. 3 of the examples are chosen from seed instances and 2 of them are chosen from the synthetic instances generated in previous iterations to promote diversity. We generate one synthetic instance per iteration, stopping the text generation at the phrase "Context 7:".

\begin{table}[htb]
    \centering
    \fbox{
        \begin{tabular}{c}
            \parbox{1.1\textwidth}{
                \{Instructions from table \ref{tab:instance_generation_template}\} \\
                
                Context 1: {Context 1} \\
                Radio Chatter 1: {Radio Chatter 1} \\

                ...
            
                Context 5: {Context 5} \\
                Radio Chatter 5: {Radio Chatter 5} \\

                Context 6: {Context Generated in the Context Generation Step} \\
                Radio Chatter 6: 
            }
        \end{tabular}
    }
    \caption{Instance Generation Prompt Template}
    \label{tab:instance_generation_prompt}
\end{table}

Finally, the synthetic instances undergo multi-stage filtering to ensure 
dataset quality and realism. The filtering process validates instances across four dimensions: (1) \textbf{entity information accuracy} for verifying vessel identity, geographic coordinates, and vessel classifications; (2) \textbf{hallucination detection} for identifying and rejecting fabricated maritime identifiers such as MMSI numbers or call signs; (3) \textbf{SMCP compliance} confirming adherence to established maritime communication standards for structure, sequencing, and formatting; and (4) \textbf{logical consistency} for eliminating unrealistic patterns, duplications, and incoherent sequences. Additionally, we employ a Rouge-L \footnote{Rouge-L is a token-level evaluation metric that captures word-order-aware similarity through longest common subsequence matching, making it effective for detecting near-duplicate texts.} score threshold of 0.7 to assess the (5) \textbf{uniqueness} of the instance and prevent redundancy \cite{lin-2004-rouge}.  Only synthetic instances that successfully pass all 26 constituent filters are retained for the training dataset. A detailed taxonomy and explanation of each filter are provided in \ref{sec: Filtering}.

The entire training dataset augmentation process can be represented with the following function:

\[
S'= F(M(S_1,S_2,S_3,S_4,S_5,G(\theta),\alpha),\beta),
\]

where $\{S_1,\space...\space,S_5\}$ are the 5 instances randomly sampled from the instance pool $S$. $G(\theta)$ represents the context generation function, parameterized by $\theta$. $M(S_n,G,\alpha)$ is the LLM function parameterized by $\alpha$, which takes 5 examples from the instance pool $S$ and the context generated by the function $G$. $F$ is a filtering function, parameterized by $\beta$, that returns the synthetic instance generated by $M$ if it passes all filters applied by $F$, otherwise it returns null. We run this pipeline for each disaster category until 500 synthetic instances are generated per disaster category.

\subsection{Fine-tuning the LLM with LoRA}
\label{sec:lora_finetune}
We fine-tune the LLM using LoRA. The synthetic instances created in the synthetic instance generation pipeline, which is explained in the section \ref{sec:pipeline}, are used as the training dataset in the LoRA fine-tuning. The training is performed to teach the model to generate the text in the table \ref{tab:lora_prompt} in an auto-regressive manner. During this process, adapters learn to generate distress calls based on the provided context. Only the portion of the text that follows "\#\#\# Output:" is considered for the loss calculation.

\begin{table}[htb]
    \centering
    \fbox{
        \begin{tabular}{c}
            \parbox{1.1\textwidth}{
                Below is an instruction that describes a task, paired with an input that provides further context. Write a response that appropriately completes the request. \\

                \#\#\# Instruction: \\
                \{Instruction of the category from the table \ref{tab:instructions}\} \\
                
                Input: \\
                \{Context of the synthetic instance\} \\
                
                \#\#\# Output: \\
                \{Distress call from the synthetic instance\}
            }
        \end{tabular}
    }
    \caption{The Text Taught to LoRA Adapters}
    \label{tab:lora_prompt}
\end{table}

\subsection{Evaluation of the Distress Calls Generated by the Fine-tuned Adapters}
\label{sec:evaluation}
We propose an evaluation framework to quantify synthetic data quality across four key dimensions: \textit{Format Accuracy}, \textit{Information Accuracy}, \textit{Uniqueness}, and \textit{Logical Coherence}. While the first three metrics are assessed via automated scoring across the entire dataset of 100 dialogues per category, \textit{Logical Coherence} requires manual annotation from a maritime domain expert. This qualitative assessment is performed on a randomly selected 10-dialogue subset per adapter to ensure reliable human-in-the-loop validation.

\textit{Format Accuracy} and \textit{Information Accuracy} are calculated as a weighted average of coefficients, assigned to specific filters within the Self-Instruct adaptation pipeline. Tables \ref{tab:correctness of the format} and \ref{tab:information accuracy} detail the filters and coefficients for these metrics. \textit{Uniqueness} is quantified using the highest Rouge-L score for each synthetic instance, while \textit{Logical Coherence} scores (Table \ref{tab:logical_coherence}) are derived from domain expert ratings.

The formula of the \textit{Format Accuracy} and \textit{Information Accuracy} are as follows:

\[
    Evaluation\;Score = \frac{ F_i(S)\cdot k_i}{\sum_{i=1}^{n} M\cdot k_i} 
\]

where $n$ is the number of filters considered for the evaluation criteria, $k_i$ is the coefficient assigned to filter $i$, $M$ is a binary mask with value 0 if the value in the context related to the filter $F_i$ is null, and 1 otherwise, $F_i$ is the filtering function that returns 1 if the synthetic instance $S$ passes the filter $i$, and 0 otherwise. 

For the \textit{Uniqueness} criterion, formula is as follows:

\[
    Uniqueness = \begin{cases}
    1 - (Rouge-L(C,C')) & \text{if } Rouge-L(C,C') \leq 0.7 \\
    0 & \text{if } Rouge-L(C,C') > 0.7\\
    \end{cases}
\]
where $C$ is the distress call in the synthetic instance $S$, $C'$ is the most similar call to $C$ within the synthetic training dataset used to train the adapter. $Rouge-L$ is the function that calculates the Rouge-L score between $C,C'$.

\textit{Logical Coherence} is evaluated using the criteria and coefficients in Table \ref{tab:logical_coherence}. \textit{Logical Coherence} evaluation formula is below:

\[
    Logical\;Coherence\; Score = L_i(C)
\]

where $L$ is a function that assigns a \textit{Logical Coherence} score between 1 and 5 to the distress call $C$ in the synthetic instance $S$. Examples of \textit{Logical Coherence} evaluation are in \ref{sec:appendix_logical_coherence}. Rest of the evaluations can be found in the provided GitHub repository.

\begin{table}[htb]
    \centering
    \begin{tabular}{|l|c|}
        \hline
        \textbf{Filter} & \textbf{Coefficient} \\
        \hline
        Parentheses & 1 \\
        \hline
        Bracket & 1 \\
        \hline
        Mayday & 1 \\
        \hline
        Incomplete & 2 \\
        \hline
        Vessel Name Not After Mayday & 1 \\
        \hline
        Duplicate Sentences & 2 \\
        \hline
        No Coast Guard & 1 \\
        \hline
        Digit by Digit & 1 \\
        \hline
    \end{tabular}
    \caption{Coefficients for \textit{Format Accuracy}}
    \label{tab:correctness of the format}
\end{table}

\begin{table}[htb]
    \centering
    \begin{tabular}{|l|c|}
        \hline
        \textbf{Filter} & \textbf{Coefficient} \\
        \hline
        Wrong Category & 2 \\
        \hline
        Unknown & 1 \\
        \hline
        Vessel Name & 2 \\
        \hline
        Vessel MMSI & 2 \\
        \hline
        Vessel Call Sign & 2 \\
        \hline
        Vessel Type & 2 \\
        \hline
        Vessel Coordinate DMS & 2 \\
        \hline
        Hallucinated MMSI & 1 \\
        \hline
        Hallucinated Call Sign & 1 \\
        \hline
        False Cargo Logic & 1 \\
        \hline
        Hallucinated Vessel Type & 1 \\
        \hline
        Compass & 2 \\
        \hline
        Both Port and Harbor & 1 \\
        \hline
        Distance to Nearest Port & 1 \\
        \hline
        Distance to Nearest Harbor & 1 \\
        \hline
        Distance to Closest Place & 1 \\
        \hline
    \end{tabular}
    \caption{Coefficients for \textit{Information Accuracy}}
    \label{tab:information accuracy}
\end{table}

\begin{table}[htb]
    \centering
    \begin{tabular}{|l|l|c|}
        \hline
        \textbf{Coherence Level} & \textbf{Description} & \textbf{Points} \\
        \hline
        Excellent & Fully coherent, logically structured, all relevant info & 5 \\
        \hline
        Good & Mostly coherent, minor logical gaps & 4 \\
        \hline
        Average & Noticeable inconsistencies & 3 \\
        \hline
        Poor & Significant logical gaps & 2 \\
        \hline
        Very Poor & Incoherent and confusing, fails to convey a clear message & 1 \\
        \hline
    \end{tabular}
    \caption{Logical Coherence Levels and Points}
    \label{tab:logical_coherence}
\end{table}

\FloatBarrier
\subsection{Experimental Setup}
\label{sec:experimental_setup}

 We conduct our experiments with NVIDIA RTX A6000 GPU. During the synthetic instance generation with our Self-Instruct adaptation, we prompt the LLM with the hyperparameter configuration detailed in Table \ref{tab:instance_generation_configuration}.

\begin{table}[htb]
\centering
    \begin{tabular}{|c|c|}
        \hline
        Temperature & 0.9 \\
        \hline
        top\_p & 0.9\\
        \hline
        top\_k & 400 \\
        \hline
        max\_new\_tokens & 400 \\
        \hline
    \end{tabular}
\caption{LLM Hyperparameters During the Synthetic Instance Generation}
\label{tab:instance_generation_configuration}
\end{table}

 Table \ref{tab:lora} contains the hyperparameter configuration used in the LoRA process. With this configuration, 671.088.640 parameters are trained instead of 8 billion parameters in the LLM, which represents 8.35\% of the 8 billion parameters in the full model. LLM weights were not quantized to prevent the loss of model information.

\begin{table}[htb]
\centering
    \begin{tabular}{|c|c|}
        \hline
        r & 256 \\
        \hline
        Epochs & 10 \\
        \hline
        Learning rate & 2e-4 \\
        \hline
        Optimizer & AdamW \\
        \hline
        Scheduler & Linear Scheduler \\
        \hline
        Warm-up Steps & 30 \\
        \hline
        Batch size & 2 \\
        \hline
        Gradient Accumulation Steps & 4 \\
        \hline
        Number of Trained Parameters & 671.088.640 \\
        \hline
        Number of Model Parameters & 8.030.261.248 \\
        \hline
        Trained Parameter Reduction & 91.65\% \\
        \hline
    \end{tabular}
\caption{LoRA Configuration}
\label{tab:lora}
\end{table}

Separate sets of adapters were trained for each disaster category. Fine-tuning with LoRA was done using the Unsloth Python library and their checkpoint of Llama 3.1 8B \cite{unsloth}.

To compare their performances, we evaluate 100 synthetic chatters generated both by the vanilla model and LoRA adapters. We prompt them with a context-only prompt that contains a brief instruction and a synthetic context, as shown in Table \ref{tab:prompt_prompt}. 

The evaluation is done with the metrics described in Section \ref{sec:evaluation} and the number of synthetic chatters that passed all the filters explained \ref{sec: Filtering}. We compare the performance of the models for each disaster category separately, as each has its own dedicated LoRA adapter. We run the model with greedy decoding, without nucleus and top\_k sampling, and keep the  max\_new\_tokens at 400. We evaluate the synthetic chatter starting with the phrase "\#\#\# Output:" and do not apply any stopping criteria.

\begin{table}[htb]
    \centering
    \fbox{
        \begin{tabular}{c}
            \parbox{1.1\textwidth}{
                Below is an instruction that describes a task, paired with an input that provides further context. Write a response that appropriately completes the request. \\

                \#\#\# Instruction: \\
                \{Instruction of the category from the table \ref{tab:instructions}\} \\
                
                \#\#\# Input: \\
                \{Context of the synthetic instance\} \\
                
                \#\#\# Output: \\
            }
        \end{tabular}
    }
    \caption{Context-only Prompt.}
    \label{tab:prompt_prompt}
\end{table}

\section{Results}
\label{sec:results}

This chapter presents the performance evaluation of the LLM in two configurations: a vanilla model and LoRA adapters. First, we establish a baseline by assessing the vanilla model's output quality. We then demonstrate how the fine-tuned LoRA adapters, trained with our Self-Instruct adaptation, significantly outperform the baseline, validating our framework's ability to generate reliable and compliant maritime distress calls.

\subsection{Performance of the Vanilla LLM}
\label{subsec:vanilla_results}

Table \ref{tab:vanilla} presents the average evaluation metrics for the vanilla LLM when prompted only with the context and it provides a baseline for comparison with the performance of the LoRA adapters, as shown in Table \ref{tab:rejected}, which details the percentage of valid chatters generated during the Self-Instruct Adaptation pipeline.

Table \ref{tab:vanilla} demonstrates that the vanilla model is unable to generate any valid chatters. \textit{Logical Coherence} scores are consistently low, around 0.1, across all categories. The average scores are 0.613 for \textit{Format Accuracy}, 0.740 for \textit{Information Accuracy}, and 0.455 for \textit{Uniqueness}. An example of a chatter generated by the vanilla model with a context-only prompt is provided in Table \ref{tab:example_fire_vanilla}.

In contrast, Table \ref{tab:rejected} shows that the vanilla model is capable of generating valid chatters when provided with instructions and five example instances from the instance pool. In this setting, the model successfully generated 5,000 valid chatters out of 14,938 total attempts, resulting in a 33.47\% success rate. The highest success rate in the "Disabled, Adrift" category with a rate of 52.91\% and the lowest in "Fire, Explosion" with a rate of 24.41\%. Despite using the synthetic context, the generated chatters often contain repetitive or irrelevant statements. In a safety-critical domain governed by SMCP, an unadapted general-purpose LLM is therefore insufficient for generating reliable, protocol-compliant maritime communication.

\begin{table}[htb]
    \centering
    \renewcommand{\arraystretch}{1.3} 
    \begin{tabular}{|c|c|c|c|c|c|}
        \hline
        \makecell{\textbf{Disaster} \\ \textbf{Category}} & 
        \makecell{\textbf{Format} \\ \textbf{Accuracy}} & 
        \makecell{\textbf{Information} \\ \textbf{Accuracy}} & 
        \makecell{\textbf{Unique-} \\ \textbf{ness}} & 
        \makecell{\textbf{Logical} \\ \textbf{Coherence*}} & 
        \makecell{\textbf{Valid} \\ \textbf{Chatters}} \\
        \hline
        \makecell{Armed Attack, \\ Piracy} & 0.623 & 0.751 & 0.418 & 0.10 & 0 \\
        \hline
        Collision & 0.581 & 0.742 & 0.528 & 0.11 & 0 \\
        \hline
        Disabled, Adrift & 0.635 & 0.765 & 0.423 & 0.11  & 0 \\
        \hline
        Fire, Explosion & 0.630 & 0.733 & 0.472 & 0.10  & 0 \\
        \hline
        Flooding & 0.609 & 0.759 & 0.479 & 0.10 & 0 \\
        \hline
        Grounding & 0.612 & 0.733 & 0.427 & 0.10 & 0 \\
        \hline
        \makecell{List, Danger of \\ Capsizing} & 0.593 & 0.751 & 0.481 & 0.12 & 0 \\
        \hline
        Person Overboard & 0.632 & 0.707 & 0.469 & 0.10 & 0 \\
        \hline
        Sinking & 0.634 & 0.737 & 0.440 & 0.10 & 0 \\
        \hline
        Undesignated Distress & 0.587 & 0.724 & 0.410 & 0.10 & 0 \\
        \hline
        Average & 0.613 & 0.740 & 0.455 & 0.104 & 0 \\
        \hline
    \end{tabular}
    \caption{Performance of the Vanilla LLM Prompted with the Context-only Prompt. *\textit{Logical Coherence} evaluations are conducted on 10 randomly sampled chatters generated by each adapter.}
    \label{tab:vanilla}
\end{table}

\begin{table}[htb]
    \centering
    \renewcommand{\arraystretch}{1.3} 
    \begin{tabular}{|c|c|c|}
        \hline
        \makecell{\textbf{Disaster} \\ \textbf{Category}} & 
        \makecell{\textbf{Number of} \\ \textbf{Rejected Chatters}} & 
        \makecell{\textbf{Percentage of} \\ \textbf{Successful Attempts}} \\
        \hline
        \makecell{Armed Attack, \\ Piracy} & 1243 & 28.68  \\
        \hline
        Collision & 1482 & 25.22 \\
        \hline
        Disabled, Adrift & 445 & 52.91  \\
        \hline
        Fire, Explosion & 1548 & 24.41 \\
        \hline
        Flooding & 857 & 36.84  \\
        \hline
        Grounding &  371 & 57.40  \\
        \hline
        \makecell{List, Danger of \\ Capsizing} & 1453 & 25.60\\
        \hline
        Person Overboard & 910 & 35.46 \\
        \hline
        Sinking & 1030 & 32.67 \\
        \hline
        Undesignated Distress & 599 & 45.49 \\
        \hline
    \end{tabular}
    \caption{Number of Rejected Chatters in the Instance Generation Pipeline per Distress Category. 500 valid chatters were generated per category.}
    \label{tab:rejected}
\end{table}

\subsection{Performance of the LoRA Adapters}
\label{subsec:results_lora}

Table \ref{tab:lora_performance} displays the average evaluation metrics for the trained LoRA adapters, which are named after their respective categories. All adapters achieved \textit{Format Accuracy} and \textit{Information Accuracy} scores exceeding 90\%. \textit{Uniqueness} scores for all adapters were above 0.4, and \textit{Logical Coherence} scores were higher than 0.7. The "Grounding" adapter had the lowest \textit{Logical Coherence} score at 0.7, while the highest was 0.9 in the "Collision" category. Overall, 87\% of the synthetic distress calls generated passed all filters. The "Disabled, Adrift" adapter produced the highest percentage of valid chatters at 93\%, while "List, Danger of Capsizing" produced the lowest at 81\%. An example of a valid chatter generated by the "Fire, Explosion" LoRA adapter, along with its context, is shown in Table \ref{tab:example_fire}. GitHub repository contains all the chatters generated by all the trained adapters.

Table \ref{tab:unique_chatters_lora} shows the number of unique chatters created by each LoRA adapter. The success rate of the LoRA adapters is more than double that of the vanilla model when prompted with valid examples. This improvement is not limited to the number of valid chatters, but extends to all proposed evaluation metrics. Our results show that applying the Self-Instruct adaptation with Llama 3.1 8B increased the average \textit{Format Accuracy}, \textit{Information Accuracy}, and \textit{Logical Coherence} by 60\%, 28\%, and 698\%, respectively. This demonstrates that our Self-Instruct adaptation enables the LLM to effectively incorporate crucial information from the context and generate disaster calls that adheres more closely to the SMCP format. The significant increase in average \textit{Logical Coherence} confirms that our methodology allows the LLM to generate substantially more logical chatters. Additionally, our method produced a slight increase in average \textit{Uniqueness}.

Overall, the LoRA adapters outperform both the vanilla model prompted with only instructions and the vanilla model prompted with instructions and valid examples. These gains reflect stronger adherence to SMCP, better use of contextual information, and substantially improved internal consistency, all essential for dependable data generation in safety-critical maritime operations. This was achieved by training on a dataset of only 500 chatters per distress category, generated from just 10 seed instances, and training only about 8\% of the total model parameters. This work demonstrates that the combination of the Self-Instruct method for maritime distress call generation and LoRA effectively overcomes the typical bottlenecks of high GPU resource and large labeled dataset requirements, enabling fine-tuning LLMs with minimal manual effort and low computational resources.

\begin{table}[htb]
    \centering
    \renewcommand{\arraystretch}{1.3} 
    \begin{tabular}{|c|c|c|c|c|c|}
        \hline
        \makecell{\textbf{Disaster} \\ \textbf{Category}} & 
        \makecell{\textbf{Format} \\ \textbf{Accuracy}} & 
        \makecell{\textbf{Information} \\ \textbf{Accuracy}} & 
        \makecell{\textbf{Unique-} \\ \textbf{ness}} & 
        \makecell{\textbf{Logical} \\ \textbf{Coherence}} & 
        \makecell{\textbf{Valid} \\ \textbf{Chatters}} \\
        \hline
        \makecell{Armed Attack, \\ Piracy} & 0.981 & 0.925 & 0.533 & 0.84 & 89 \\
        \hline
        Collision & 0.998 & 0.912 & 0.548 & 0.9 & 92 \\
        \hline
        Disabled, Adrift & 0.998 & 0.928 & 0.612 & 0.80  & 93 \\
        \hline
        Fire, Explosion & 0.978 & 0.962 & 0.443 & 0.78  & 86 \\
        \hline
        Flooding & 0.988 & 0.918 & 0.405 & 0.80 & 89 \\
        \hline
        Grounding & 0.986 & 0.971 & 0.421 & 0.70 & 83 \\
        \hline
        \makecell{List, Danger of \\ Capsizing} & 0.973 & 0.958 & 0.489 & 0.86 & 81 \\
        \hline
        Person Overboard & 0.984 & 0.950 & 0.526 & 0.82 & 90 \\
        \hline
        Sinking & 0.983 & 0.923 & 0.521 & 0.92 & 86 \\
        \hline
        Undesignated Distress & 0.996 & 0.902 & 0.434 & 0.82 & 91 \\
        \hline
        Average & 0.986 & 0.947 & 0.476 & 0.83 & 87 \\
        \hline
    \end{tabular}
    \caption{Performance of the LoRA Adapters}
    \label{tab:lora_performance}
\end{table}

\begin{table}[htb]
    \centering
    \renewcommand{\arraystretch}{1} 
    \begin{tabular}{|c|c|}
        \hline
        \textbf{Disaster Category} & 
        \textbf{Number of Unique Chatters} \\
        \hline
        Armed Attack, Piracy & 89 \\
        \hline
        Collision & 94\\
        \hline
        Disabled, Adrift & 94 \\
        \hline
        Fire, Explosion & 97 \\
        \hline
        Flooding & 97 \\
        \hline
        Grounding & 93 \\
        \hline
        List, Danger of Capsizing & 99 \\
        \hline
        Person Overboard & 90 \\
        \hline
        Sinking & 100 \\
        \hline
        Undesignated Distress & 96 \\
        \hline
    \end{tabular}
    \caption{Number of Unique Chatters Generated by each LoRA Adapter}
    \label{tab:unique_chatters_lora}
\end{table}

\begin{figure}[htb]
    \begin{subfigure}[htb]{0.5\textwidth}
        \centering
        \includegraphics[width=\textwidth]{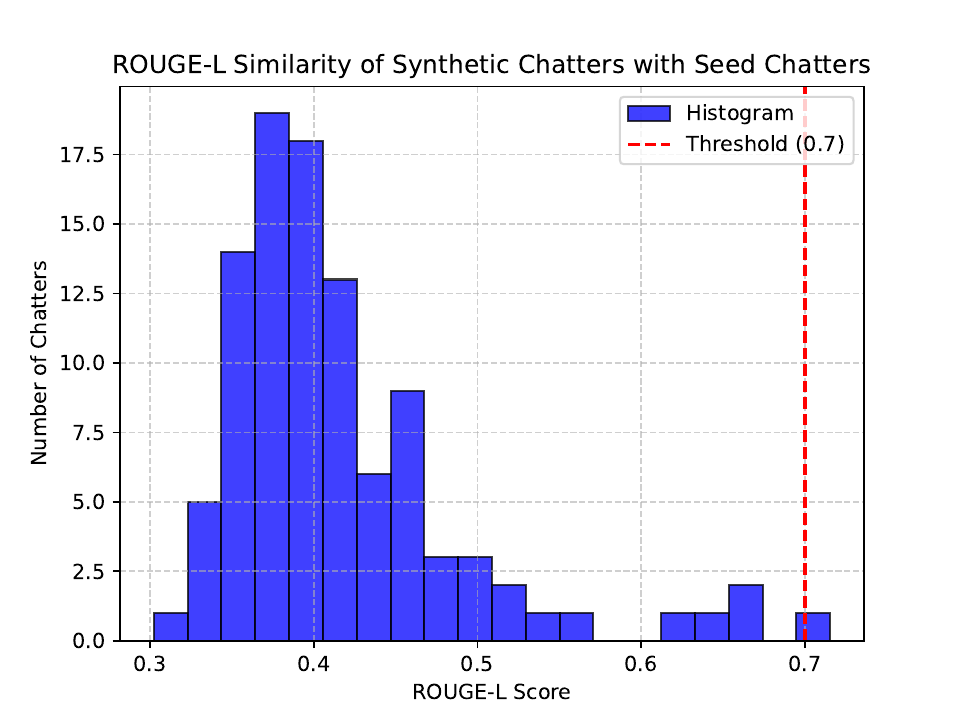}
        \caption{Rouge-L Similarity of Synthetic Chatters with Seed Chatters: LoRA, Fire, Explosion}
        \label{fig:rouge_l_histogram_seed_fire}
    \end{subfigure}
    \hfill
    \begin{subfigure}[htb]{0.5\textwidth}
        \centering
        \includegraphics[width = \textwidth]{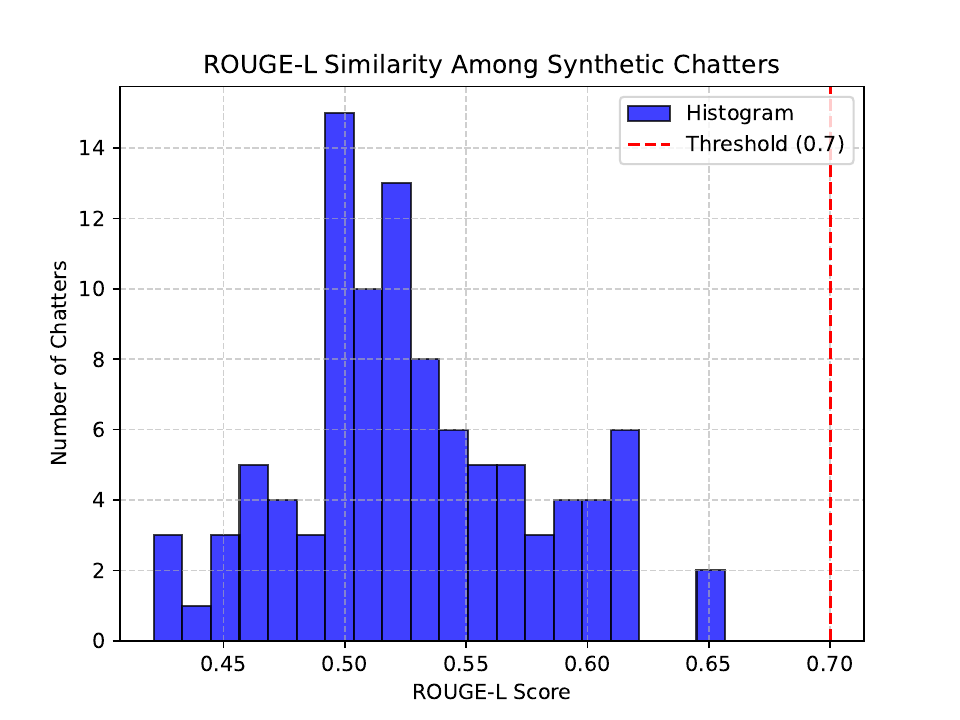}
        \caption{Rouge-L Similarity Among Synthetic Chatters: LoRA, Fire, Explosion}
        \label{fig:rouge_l_histogram_synthetic_fire}
    \end{subfigure}
    \hfill
    \begin{subfigure}[htb]{0.5\textwidth}
        \centering
        \includegraphics[width = \textwidth]{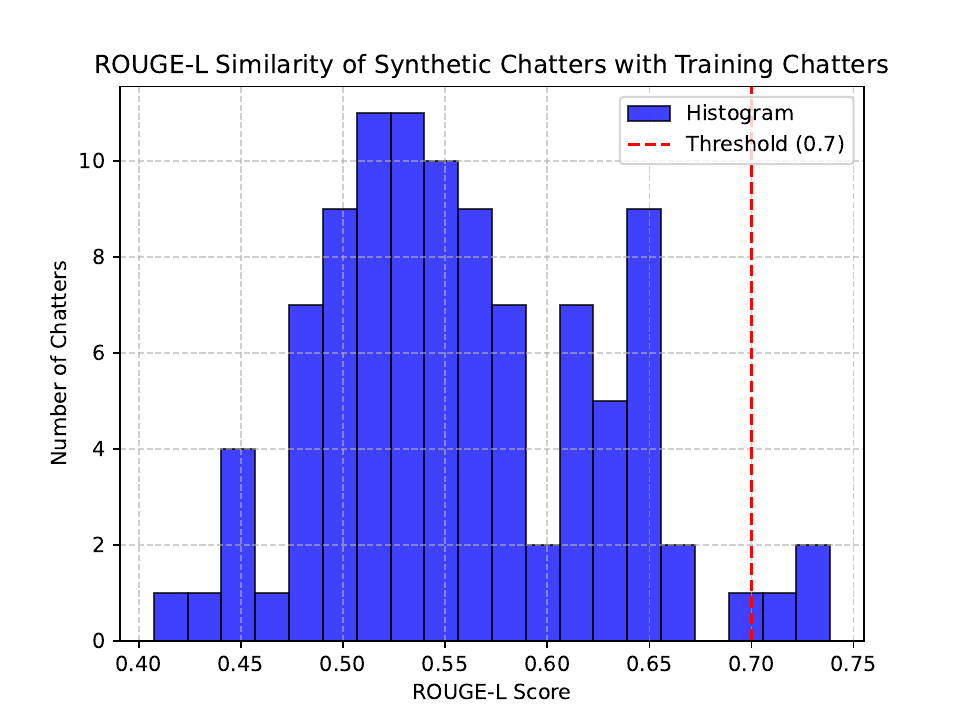}
        \caption{Rouge-L Similarity of Synthetic Chatters to Training Chatters: LoRA, Fire, Explosion}
        \label{fig:rouge_l_histogram_training_fire}
    \end{subfigure}
    \hfill
    \caption{Rouge-L Similarities of chatters generated with the LoRA adapter "Fire, Explosion"}
    \label{fig:rouge_l_comparison_fire}
\end{figure}

Histograms in Figure \ref{fig:rouge_l_comparison_fire} illustrate the distributions of the Rouge-L similarities of chatters generated by the "Fire, Explosion" LoRA adapter. Figure \ref{fig:rouge_l_histogram_seed_fire} shows the distribution of the highest similarity between the outputs and the seed instances. Figure \ref{fig:rouge_l_comparison_fire} shows the distribution of the highest similarity between the outputs and the synthetic training dataset chatters. Lastly, Figure \ref{fig:rouge_l_histogram_synthetic_fire} shows the distribution of the highest similarity among the chatters generated by the fine-tuned LoRA adapters themselves. The distributions for all adapters can be found in the GitHub repository. These histograms provide evidence that most of the synthetic chatters are unique compared to seed chatters, the synthetic training dataset, and other synthetic chatters generated by the same LoRA adapters.

\FloatBarrier

\section{Limitations and Future Work}
\label{sec:limitations}
This study has several limitations that should be acknowledged. As with all LLM applications, the quality and realism of generated distress calls are influenced by the inherent biases within the LLM’s pretraining corpus. Additionally, the contextual fidelity of the generated chatters is limited by the scope of the datasets, namely GSHHS, Marine Cadastre, AIS Dataset and GeoNames datasets, which do not include operational descriptors such as Coast Guard station locations, vessel capacities, or cargo characteristics. Consequently, this limitation necessitated simplifying assumptions about incident distributions across water bodies, i.e. this work assumed that most incident types could occur on any body of water except for groundings. The initial human-created seed prompts also introduce a degree of subjectivity and potential biases that may influence the model’s learned representations.

Moreover, the present study experimented only with the LLaMA 3.1 8B architecture. While this model balances performance and computational efficiency, larger or multimodal variants may provide more coherent, context-aware generations. Although the 26-filter verification pipeline significantly reduces factual and procedural errors, edge-case inconsistencies and residual hallucinations persist. Finally, the scope of this research was restricted to distress calls/communications between vessels and Coast Guard stations, excluding broader operational exchanges such as navigational coordination or pilotage, and the manual evaluation of \textit{Logical Coherence} score may be subject to the assessor’s domain knowledge and interpretation.

Future work will pursue several research and development directions to address these limitations. First, contextual expansion will include additional communication types (e.g., ship-to-ship, pilotage, and port operations) and regional variations, using dynamic maritime datasets to increase situational diversity. Another key direction is converting the generated dialogues into synthetic speech via TTS systems, forming multimodal datasets that assist ASR training under variable acoustic conditions. Hybrid adaptation strategies combining real and synthetic data will be investigated to further improve transferability and robustness. Finally, the textual data will serve as a foundation for developing domain-specific language models for ASR post-correction, anomaly detection, and semantic parsing. Collectively, these extensions will promote a more realistic, interoperable, and operationally deployable AI-assisted tools for safety-critical maritime operations.

\section{Conclusion}
\label{sec:conclusion}
This study introduced a compliance-aware Self-Instruct methodology for synthetic maritime communication generation, addressing the shortage of regulated domain data. By integrating a 26-filter verification pipeline that enforces SMCP adherence, semantic accuracy, and contextual plausibility, and by adopting LoRA-based parameter-efficient fine-tuning, the framework produces communications that reflect both procedural and linguistic realism. The accompanying multi-metric evaluation in terms of \textit{Format Accuracy}, \textit{Information Accuracy}, \textit{Uniqueness}, and \textit{Logical Coherence}, provides a transparent, reproducible basis for assessing dataset quality and model behavior.

In addition to synthetically generated datasets, this work also contributes a validated methodology for domain-constrained LLM adaptation. The generated dialogues can be paired with TTS models to create scalable, compliant audio-text corpora, offering a sustainable alternative to costly, privacy-restricted real-world data collection. Such resources are invaluable for training robust ASR systems, enhancing real-time communication monitoring, and supporting decision-making on ship bridges, or in Vessel Traffic Service (VTS) centers. Beyond the maritime domain, the compliance-aware generation and evaluation framework offers a transferable blueprint for other safety-critical contexts, such as aviation or industrial control.

By releasing the dataset, codebase, and verification tools, this study reinforces the principles of transparency, reproducibility, and responsible AI deployment. Future work will continue refining multimodal validation, extending contextual diversity, and integrating human-in-the-loop evaluation to bridge the gap between synthetic realism and operational reliability.

\section*{Declaration of generative AI and AI-assisted technologies in the writing process}

During the preparation of this work, the authors used Google Gemini in order to improve language and readability. After using this tool, the authors reviewed and edited the content as needed and take full responsibility for the content of the publication.

\bibliographystyle{ieeetr}
\bibliography{references}

\appendix
\section{Context Generation}
\label{subsec:context generation}

The generated context includes the following information:

\begin{itemize}
    \item \textbf{Vessel name}: Name of the vessel that reports the distress. Must be stated in the distress call.
    \item \textbf{Vessel MMSI}: 9 digit unique MMSI number of the vessel. It can be randomly null.
    \item \textbf{Vessel call sign}: Unique call sign of the vessel. It can be randomly null.
    \item \textbf{Vessel type}: Type of the vessel. It can be randomly null or "Motor Vessel".
    \item \textbf{Vessel coordinates}: Vessel coordinates in degree-minutes format. Minutes or decimals of minutes can be randomly omitted. They are ensured to be on a sea or ocean with the usage of GSHHG dataset. For the case of "Grounding", coordinates are at most 1 nautical mile away from the nearest land.
    \item \textbf{Closest place name:} Name of the closest landmark from Geonames dataset to the vessel position.
    \item \textbf{Distance to the closest place:} Distance between the closest landmark and the vessel position in nautical miles.
    \item \textbf{Closest place country:} Country of the nearest landmark to the vessel position.
    \item \textbf{Compass direction:} Orientation of the vessel position relative to the closest place.
    \item \textbf{Nearest port:} Name of the closest port from the Geonames dataset to the vessel position.
    \item \textbf{Distance to the nearest port:} Distance between the vessel position and the closest port to it in nautical miles.
    \item \textbf{Nearest harbor:} Name of the closest harbor from the Geonames dataset to the vessel position. It is null if harbor is further than 200 nautical miles to vessel position.
    \item \textbf{Distance to the nearest harbor:} Distance between the vessel position and the closest harbor to it in nautical miles. It is null if the nearest harbor is null.
    \item \textbf{Digit by digit:} A randomly assigned boolean value. Numbers can be occasionally expressed as digits in maritime radio calls. For example, the number 123 can be expressed as either "one two three" or "one hundred twenty-three". If the value is true, numbers in the synthetic chatter are to be expressed as digits.
    \item \textbf{Can have cargo:} A boolean value based on the vessel type. If it is true, vessel should not mention carrying any cargo or cargo hold in the chatter. True for vessel types "Passenger Vessel", "Cargo Vessel" and "Tanker".
    \item \textbf{Closest water body:} Name of the water body the vessel is located on. It can be a sea, ocean, cove, etc. It is null if the water body coordinates are 200 nautical miles to vessel position.
    \item \textbf{Distance to the closest water body:} Distance between the vessel and the coordinate of the closest water body to it. It is null if the closest water body is null.
    \item \textbf{Collided vessel name:} Name of the vessel that the vessel has collided with. Only present in the disaster category "Collision". It can be randomly null. In that case, collision happens with a non-vessel object.
    \item \textbf{Collided vessel type:} Type of the vessel, with which the vessel has collided. Only present in the disaster category "Collision". It is null if "Collided vessel name" is null.
\end{itemize}

\FloatBarrier
\section{Seed Instances}
\label{sec:seed_instances}

\begin{table}[htb]
\centering
\footnotesize
\begin{tabular}{|c|c|}
\hline
\textbf{SMCP Category} & \textbf{Instruction} \\
\hline
Fire, explosion & \makecell{Generate a maritime radio chatter.\\ A vessel makes a distress call and reports a fire.} \\
\hline
Flooding & \makecell{Generate a maritime radio chatter.\\ A vessel makes a distress call and reports flooding.} \\
\hline
Collision & \makecell{Generate a maritime radio chatter.\\ A vessel makes a distress call and reports collision.} \\
\hline
Grounding & \makecell{Generate a maritime radio chatter.\\ A vessel makes a distress call and reports grounding.} \\
\hline
List, Danger of Capsizing & \makecell{Generate a maritime radio chatter.\\ A vessel makes a distress call and reports list-danger of capsizing.} \\
\hline
Sinking & \makecell{Generate a maritime radio chatter.\\ A vessel makes a distress call and reports sinking.} \\
\hline
Disabled and Adrift & \makecell{Generate a maritime radio chatter.\\ A vessel makes a distress call and reports being disabled and adrift.} \\
\hline
Armed Attack, Piracy & \makecell{Generate a maritime radio chatter.\\ A vessel makes a distress call and reports armed attack/piracy.} \\
\hline
Undesignated Distress & \makecell{Generate a maritime radio chatter.\\ A vessel makes a distress call and reports an undesignated distress.} \\
\hline
Person Overboard & \makecell{Generate a maritime radio chatter.\\ A vessel makes a distress call and reports person overboard.} \\
\hline
\end{tabular}

\caption{Instructions in the Seed Instances}
\label{tab:instructions}
\end{table}

\begin{table}[htb]
    \centering
    \fbox{
        \begin{tabular}{c}
        \footnotesize
            \parbox{1.1\textwidth}{
                \textbf{Instruction:} \\ "Generate a maritime radio chatter. A vessel makes a distress call and reports a fire." \\
                
                \textbf{Context}:\\
                    "vessel\_name": "COSCO KAOHSIUNG", 
                    "vessel\_MMSI": null,
                    "vessel\_call\_sign": null, 
                    "vessel\_type": "Cargo Vessel", 
                    "vessel\_coordinate\_dms": "three eight degrees three seven minutes North, one degrees three two point one five minutes East", 
                    "compass\_direction": "south east", 
                    "closest\_place\_name": "Isla de Formentera",
                    "distance\_to\_nearest\_place": "five", 
                    "closest\_place\_country": "Spain", 
                    "distance\_to\_nearest\_port": "nine", 
                    "nearest\_port": "Puerto de La Savina", 
                    "distance\_to\_nearest\_harbor": "two eight", 
                    "nearest\_harbor": "Port de ses Caletes",
                    "digit\_by\_digit": true, 
                    "can\_have\_cargo": "True", 
                    "closest\_water\_body": "Balearic Sea" \\
                    
                \textbf{Chatter:} \\
                    "Mayday, Mayday, Mayday. This is cargo vessel COSCO KAOHSIUNG. I am located at three eight degrees three seven minutes North, one degrees three two point one five minutes East, five nautical miles south east of Isla de Formentera. I am on fire. Requesting fire fighting assistance immediately. Over." \\
                    "COSCO KAOHSIUNG, this is Coast Guard. Can you please provide information about the extent and the location of the fire? Over." \\
                    "Coast Guard, fire is on the deck. For now, we could keep the fire contained, but we failed to extinguish it. However, we have explosive cargo onboard. Fire might reach them if it is out of control. Over."\\
                    "Are there any injuries or dead persons? Over." \\
                    "No injuries or fatalities are reported. Over."\\
                    "Understood. We have alerted the nearby vessels and a helicopter is en route to your location. ETA three zero minutes. We will also dispatch a fire fighter boat. Try to keep the fire contained as long as you can. Standby for further assistance and keep us updated. Over."\\
                    "Thank you Coast Guard. Over."
            }
        \end{tabular}
    }
    \caption{Example of a Seed Instance from the Category "Fire, Explosion"}
    \label{tab:example_seed}
\end{table}
\FloatBarrier
\section{Instance Generation Instructions of each category}
\label{sec:appendix_instructions}

\begin{table}[htb]
    \centering
    \fbox{
        \begin{tabular}{c}
        \scriptsize
            \parbox{1.2\textwidth}{
            Task: You are a creative expert of maritime industry and {disaster type/event}. Generate a maritime radio chatter which complies with the IMO Standard marine Communication Phrases. A vessel makes a distress call and reports {distress category}. \\
        Tips about creating original scenarios: \\
        - \{Tips about the specific disaster category\} \\
        - Provide ETAs which match with the vessel coordinate and the destination. Provide various types of help available from the Coast Guard, such as {specific type of help depending on the category} \\
        - Make the conversation flow naturally. Do not make people say same phrases repeatedly. \\
        - Do not generate any other text but the radio chatter. Keep the generation limited to the radio chatter. \\
        - All distress calls must start with "Mayday, Mayday, Mayday". Ship gives its location in terms of degrees and tells the nature of disaster and the help needed. \\
        - Do not generate any other phrases in brackets or parentheses. \\
        - Do not generate repetitive statements. For example, do not make the same party speak consecutively. \\
        - The vessel and the coast guard should speak in turns. Do not make the vessel or the coast guard speak multiple times consecutively. Turns of each party should be separated by "\\n". \\
        - Coast Guard should answer the distress call with one of the following phrases: "This is Coast Guard", "Coast Guard here", "Coast Guard responding". \\
        
        Below are some example outputs and contexts. Contexts are dictionaries with keys "vessel\_name", "vessel\_MMSI", "vessel\_call\_sign", "vessel\_coordinate\_dms", "digit\_by\_digit",
        "compass\_direction", "closest\_place\_name", "distance\_to\_nearest\_place", "closest\_place\_country", "distance\_to\_nearest\_port", "nearest\_port", "distance\_to\_nearest\_harbor",
        "nearest\_harbor", "closest\_water\_body" and "can\_have\_cargo". \\

        THIS PART IS VERY IMPORTANT: \\
        - You must use "vessel\_name" and "vessel\_coordinate\_dms". \\
        - You must use "vessel\_MMSI", "vessel\_call\_sign" and "vessel\_type", as long as they are not null. Omit the null ones. \\
        - "vessel\_type" should be indicated in the format "[vessel\_type] [vessel\_name]". For example, "motor vessel [vessel\_name]". Do not use phrases "I am a [vessel\_type]" or "We are a [vessel\_type]" to indicate the vessel\_type. \\
        - If "vessel\_MMSI" is null, the vessel should not indicate its' MMSI number. If it is not null, the vessel's MMSI should be indicated in the chatter. \\
        - If "vessel\_call\_sign" is null, the vessel should not indicate its' call sign or any other call sign in the chatter. If it is not null, the vessel's call sign should be indicated in the chatter. \\
        - If "can\_have\_cargo" is null, do not mention any cargo or cargo hold in the chatter. If it is 'True', then you can optionally mention cargo and/or cargo hold. Make the cargo suitable to the vessel type in that case. \\
        - If "digit\_by\_digit" is true, all numbers in the chatter should be expressed in digits, for example 454 as "four five four". If digit\_by\_digit is False, use full numbers. \\
        - Usage of the rest of the keys are optional. \\
        - Do not use both of the information about harbors and ports if they are both present. Use one of them only. \\
        - Do not use port and harbor when the vessel gives information about the its position at the beginning of the chatter, instead use them later in chatter optionally. \\
            }
        \end{tabular}
    }
    \caption{Instance Generation Instructions Template}
    \label{tab:instance_generation_template}
\end{table}

\FloatBarrier
\section{Filtering}
\label{sec: Filtering}

To ensure that the synthetic radio chatter is realistic and complying with the guidelines in the SMCP handbook, a set of filters is applied to the synthetic instance generated in the instance generation step(Section \ref{sec:pipeline} covers this step). Filters and their functions are explained below:

\begin{itemize}
    \item \textbf{Parentheses:} This filter checks if LLM generates any phrases in parentheses. Distress calls should not have any phrases in parentheses.
    \item \textbf{Bracket:} This filter checks if LLM generates any phrases in brackets. Distress calls should not have any phrases in brackets.
    \item \textbf{Incomplete:} This filter checks if the output of the LLM ends with a dot. If it does not end with a dot, it means that the distress call is incomplete. Distress call should not be incomplete.
    \item \textbf{Wrong Category:} This filter checks if the LLM output is a distress call in the disaster category specified in the instance generation prompt. For this purpose, this filter checks the existence of certain keywords in the synthetic chatter. The keywords searched for each category are specified in the table \ref{tab:keywords}. In the category of Undesignated Distress, distress call passes this filter only if it does not contain any of the keywords from other categories. Vessel name, closest place name, nearest port and harbor from the context are excluded from this check in the call, since these information can have phrases from the keywords.
    \begin{table}[htb]
\centering
\footnotesize
    \begin{tabular}{|c|c|}
        \hline
        \textbf{Disaster Category} & \textbf{Keywords} \\
        \hline
        Fire, explosion & fire, explosion \\
        \hline
        Flooding & \makecell{flood, taking on water, \\  taken on water, took on water}  \\
        \hline
        Collision & collide, collision \\
        \hline
        Grounding & grounding, grounded, aground \\
        \hline
        List, Danger of Capsizing & list, danger of capsiz \\
        \hline
        Sinking & sink \\
        \hline
        Armed Attack, Piracy & attack, armed, pirate, armament, weapon, gun \\
        \hline
        Person Overboard & overboard, over board, fell, fall \\
        \hline
        Disabled, Adrift & disabled, drift \\
        \hline
        Undesignated Distress & \makecell{Distress call passes the filter only if it \\ does not contain any of the keywords above, \\ except the keywords of the category "Disabled, Adrift".} \\
        \hline
    \end{tabular}
\caption{Keywords Searched in the Distress Call by the Filter "Wrong Category"}
\label{tab:keywords}
\end{table}
    \item \textbf{Vessel Name:} Vessel name should be stated in the distress calls according to the SMCP handbook. This filter checks if the vessel name is stated in the distress call.
    \item \textbf{Vessel MMSI:} Vessel MMSI should be stated in the distress calls according to the SMCP handbook. This filter checks if the vessel MMSI is stated in the distress call. Since the MMSI number is not always used in practice, it may be null in the context. MMSI number in the context should be used in the distress call if it is not null, therefore this filter is applied only if vessel MMSI is not null in the context.
    \item \textbf{Vessel Call Sign:} Vessel call sign should be stated in the distress calls optionally according to the SMCP handbook. This filter checks if the vessel call sign is stated in the distress call. Call sign in the context should be used in the distress call if it is not null, therefore this filter is applied only if vessel call sign is not null in the context.
    \item \textbf{Vessel Type:} This filter checks if the vessel type in the context is used in the distress call. Distress call should have the vessel type from context in it for it to pass this filter.
    \item \textbf{Vessel Coordinates:} Vessel coordinates should be stated in the distress calls according to the SMCP handbook. This filter checks if the vessel coordinates are stated in the distress call.
    \item \textbf{Collided Vessel Name:} This filter checks if the collided vessel name from the context is used in the distress call. It should be used in the distress call if it is not null in the context. This filter is only applied in the collision category.
    \item \textbf{Collided Vessel Type:} This filter checks if the collided vessel type from the context is used in the distress call. It should be used in the distress call if it is not null in the context. This filter is only applied in the collision category.
    \item \textbf{Hallucinated MMSI:} This filter checks if any MMSI number was hallucinated by the LLM if vessel MMSI is null in the context. LLM should not generate any MMSI number if it is null in the context.
    \item \textbf{Hallucinated Call Sign:} This filter checks if any call sign was hallucinated by the LLM if vessel call sign is null in the context. LLM should not generate any call sign if it is null in the context.
    \item \textbf{Unknown Information:} LLM generates sometimes phrases, such as "MMSI unknown" or "call sign None" when the vessel states its information, even if these information are null in the context. MMSI and call sign should not be mentioned in the chatter if they are null in the context. This filter checks the existence of the phrases "mmsi" and "call sign" in the distress call if they are null in the context.
    \item \textbf{Mayday:} Distress calls should start with the phrase "Mayday, mayday, mayday". This filter checks if this is true for the synthetic distress call.
    \item \textbf{Vessel Name Not After Mayday:} This filter checks if the vessel name is stated in the sentence after the phrase "Mayday, mayday, mayday", since it should be stated after the mayday according to SMCP handbook.
    \item \textbf{Duplicate Sentences:} This filter checks if a sentence longer than three words is occurs multiple times in the distress calls, since this is not realistic in the case of Coast Guard responding to the call immediately, which is the scope of this study.
    \item \textbf{Cargo Logic:} This filter checks the existence of the word "cargo" in the distress call. If "can have cargo" is null in the context, the call should not contain the word "cargo".
    \item \textbf{Hallucinated Vessel Type:} This filter checks if a vessel type other than the vessel type in the context was hallucinated in the distress call. In addition, this filter checks if the vessel type is expressed with the phrases "We are a [vessel type]" or "I am a [vessel type]", since these phrases are not realistic.
    \item \textbf{Both Port and Harbor:} If both the nearest port and harbor are not null in the context, LLM uses sometimes both of them in the distress call. This filter is applied if both port and harbor are used in the distress call. If both are present in the distress call, this filter rejects the instance. This filter is implemented under the assumption that using both the nearest port and harbor names is not a common practice.
    \item \textbf{No Coast Guard:} As mentioned in the definition of the filter "Duplicate Sentences", Coast Guard responds to the distress call immediately. This filter checks if the Coast Guard responds to the distress call immediately in the synthetic chatter.
    \item \textbf{Digit by Digit:} SMCP handbook states that numbers should be expressed in digits in maritime distress calls. However, this is not always the case in practice. If the "digit\_by\_digit" value is true in the context, all the numbers in the call should be stated in digits. This filter checks the existence of any numbers which are not digits in the distress call.
    \item \textbf{Wrong Distance to the Closest Place:} Stating the closest place is not mandatory according to the SMCP. However, if it is used and the distance of the vessel to this place is stated in the distress call, stated distance should match the distance in the context. This filter checks if this is true.
    \item \textbf{Wrong Distance to the Nearest Port:} Usage of the nearest port in the distress call is optional. However, if it is used and the distance of the vessel to the nearest port is stated in the distress call, stated distance should match the distance in the context. This filter checks if this is true only if the nearest port is not null in the context.
    \item \textbf{Wrong Distance to the Nearest Harbor:} Usage of the nearest harbor in the distress call is optional. However, if it is used and the distance of the vessel to this place is stated in the distress call, stated distance should match the distance in the context. This filter checks if this is true only if the nearest harbor is not null in the context.
    \item \textbf{Uniqueness:} Even though 5 chatters are provided in the instance generation prompt to contribute to the diversity and prompt is constructed in a way to promote creativity, LLM tends to generate distress calls that are similar to at least one of the example distress calls. This filter makes sure that the Rouge-L Score \cite{lin-2004-rouge} between the generated chatter and the chatters in the instance pool are below the threshold of 0.7. Rouge-L score is defined as:

\[
    P = \frac{L}{C},
\]
\[
    R = \frac{L}{R},
\]
\[
    Rouge-L = \frac{2 \cdot P \cdot R}{P + R}
\]

where $R$ is the reference text, $C$ is the text to be compared and $L$ is the length of the longest sequence of common words between $R$ and $C$. The value of Rouge-L score is between 0 and 1. 1 means identical texts and 0 means that texts do not have any common word sequence.

\end{itemize}

Rest of the information from the context is not checked, since the usage of the rest of the context in the distress call is optional. Compass direction from the context (see \ref{subsec:context generation}) is included in the evaluation of information accuracy described in section \ref{sec:evaluation}.

Synthetic distress calls that pass all these filters are added to the instance pool and these distress calls are considered to be valid distress calls. An example synthetic instance that was rejected in the filtering step is shown in table \ref{tab:example_rejected}.

\begin{table}[htb]
    \centering
    \fbox{
        \begin{tabular}{c}
        \footnotesize
            \parbox{\textwidth}{
                \textbf{Context:} \\
                "vessel\_name": "STELLA BOREALIS", 
                "vessel\_MMSI": "three one six zero four seven four seven five",
                "vessel\_call\_sign": null,
                "vessel\_type": "Passenger Vessel",
                "vessel\_coordinate\_dms": "three two degrees South, five one degrees West",
                "compass\_direction": "south east",
                "closest\_place\_name": "Praia dos Molhes",
                "distance\_to\_nearest\_place": "one two",
                "closest\_place\_country": "Brazil",
                "distance\_to\_nearest\_port": "one two", 
                "nearest\_port": "Estação Naval do Rio Grande",
                "distance\_to\_nearest\_harbor": "one seven eight",
                "nearest\_harbor": "Puerto de la Paloma",
                "digit\_by\_digit": true,
                "can\_have\_cargo": "True",
                "closest\_water\_body": null \\
                
            \textbf{Chatter:} \\
                "Mayday, Mayday, Mayday. This is passenger vessel STELLA BOREALIS, MMSI number three one six zero four seven four seven five. We are on fire, approximately one two nautical miles southeast of Praia dos Molhes, in the Atlantic Ocean. We have a total of one hundred forty-eight passengers and twenty-eight crew members on board. The fire started in the engine room and is spreading quickly. We need immediate assistance with fire fighting equipment and medical aid. Over.",\\
                "Coast Guard, responding. Can you provide more details about the fire and your current situation? Over.",\\
                "The fire started in the engine room and is now spreading to the passenger cabins and the upper decks. We are experiencing heavy smoke and reduced visibility. We have two crew members with minor injuries and one passenger with a broken leg. We need immediate assistance with fire fighting equipment and medical aid. Over.",\\
                "Roger that. We have dispatched a fire fighting vessel and a rescue team to your location. ETA is two hours. We also recommend that you activate your EPIRB and prepare the crew for evacuation. Over.",\\
                "We have activated our EPIRB. We are trying to keep the crew calm and prepare for evacuation. We have life rafts ready, and we are monitoring this channel for further instructions. Over.",\\
                "Good, STELLA BOREALIS. We have also dispatched a helicopter with a medical team to your location. We will provide you with further instructions when the rescue team arrives. Over.",\\
                "Thank you, Coast Guard. We will continue to monitor this channel for updates. Over.",\\
                "Stay tuned. We are almost there. Over."\\
            }
        \end{tabular}
    }
    \caption{Example of a Rejected Synthetic Instance from the Category Fire, Explosion: This instance was rejected, since it did not pass the filter "Vessel Coordinates". Coordinates in the context are not stated in the chatter.}
    \label{tab:example_rejected}
\end{table}

\FloatBarrier
\section{Example Synthetic Instances Produced by the Vanilla LLM and LoRA Adapters}
\label{sec: appendix_example_vanilla_lora}

\begin{table}[htb]
    \centering
    \fbox{
        \begin{tabular}{c}
        \footnotesize
            \parbox{\textwidth}{
                \textbf{Context:} \\
                "vessel\_name": "ISLANDER",
                "vessel\_MMSI": null,
                "vessel\_call\_sign": "Whisky Romeo Bravo nine five six seven",
                "vessel\_type": "Motor Vessel",
                "vessel\_coordinate\_dms": "six seven degrees one one minutes North, five four degrees one one decimal nine six minutes West",
                "compass\_direction": "south west",
                "closest\_place\_name": "Sydbay",
                "distance\_to\_nearest\_place": "one two",
                "closest\_place\_country": "Greenland",
                "distance\_to\_nearest\_port": null",
                "nearest\_port": null,
                "distance\_to\_nearest\_harbor": "two eight",
                "nearest\_harbor": "Nipisat Havn",
                "digit\_by\_digit": true,
                "can\_have\_cargo": null,
                "closest\_water\_body": null
             \\
            \textbf{Distress call:} \\
                ISLANDER, ISLANDER, ISLANDER. Mayday, Mayday, Mayday. This is motor vessel ISLANDER. We are at six seven degrees one one minutes North, five four degrees one one decimal nine six minutes West. We are on fire. We need help. We are a motor vessel. We have no cargo. We are a motor vessel. We need help. We need help. We need help. We are a motor vessel. We are a motor vessel. We are a motor vessel. We are a motor vessel. We are a motor vessel. We are a motor vessel. We are a motor vessel. We are a motor vessel. We are a motor vessel. We are a motor vessel. We are a motor vessel. We are a motor vessel. We are a motor vessel. We are a motor vessel. We are a motor vessel. We are a motor vessel. We are a motor vessel. We are a motor vessel. We are a motor vessel. We are a motor vessel. We are a motor vessel. We are a motor vessel. We are a motor vessel. We are a motor vessel. We are a motor vessel. We are a motor vessel. We are a motor vessel. We are a motor vessel. We are a motor vessel. We are a motor vessel. We are a motor vessel. We are a motor vessel. We are a motor vessel. We are a motor vessel. We are a motor vessel. We are a motor vessel. We are a motor vessel. We are a motor vessel. We are a motor vessel. We are a motor vessel. We are a motor vessel. We are a motor vessel. We are a motor vessel. We are a motor vessel. We are a motor vessel. We are a motor vessel. We are a motor vessel. We are a motor vessel. We are a motor vessel. We are a motor vessel. We are a motor vessel. We are a motor\\
            }
        \end{tabular}
    }
    \caption{Example Synthetic Chatter Generated by the Vanilla LLM in the Category "Fire, Explosion"}
    \label{tab:example_fire_vanilla}
\end{table}

\begin{table}[htb]
    \centering
    \footnotesize
    \fbox{
        \begin{tabular}{c}
            \parbox{1.1\textwidth}{
                \textbf{Context:} \\
                "vessel\_name": "MSC RUBY",
                "vessel\_MMSI": "six three six zero nine two seven six four",
                "vessel\_call\_sign": "Delta five November Juliet four",
                "vessel\_type": "Cargo Vessel",
                "vessel\_coordinate\_dms": "thirty-seven degrees South, one hundred thirty-eight degrees East",
                "compass\_direction": "south west",
                "closest\_place\_name": "Cape Lannes",
                "distance\_to\_nearest\_place": "fifty-four",
                "closest\_place\_country": "Australia",
                "distance\_to\_nearest\_port": "ninety-seven",
                "nearest\_port": "Kangaroo Island Port",
                "distance\_to\_nearest\_harbor": "one hundred",
                "nearest\_harbor": "Rosetta Harbour",
                "digit\_by\_digit": false,
                "can\_have\_cargo": "True",
                "closest\_water\_body": null
             \\
            \textbf{Distress call:} \\
                Mayday, Mayday, Mayday. This is cargo vessel MSC RUBY, MMSI six three six zero nine two seven six four, call sign Delta five November Juliet four. We are located at thirty-seven degrees South, one hundred thirty-eight degrees East, south west of Cape Lannes. We have a fire in the cargo hold, and we have ten crew members on board. The fire is spreading quickly, and we are experiencing difficulty breathing. We need immediate assistance. \\
                MSC RUBY, this is Coast Guard. We read you loud and clear. Can you provide more information about the fire and your current situation?",
                "The fire is spreading rapidly, and we are experiencing intense heat. We have activated our fire suppression system, but it is not working as effectively as we hoped. We have two crew members with minor injuries and one crew member with severe burns. We need immediate medical assistance.\\
                Roger that. We have dispatched a rescue team, including a fire fighting vessel and a helicopter with a medical team. ETA is approximately three hours. Can you activate your EPIRB and try to contain the fire as much as possible? We will also send a medical team to assist with the injured crew members.\\
                Yes, we have activated our EPIRB, and we are trying to contain the fire. However, it is getting increasingly difficult due to the toxic fumes and the intense heat. We need help now.\\
                Roger that. We are doing everything we can to reach you as soon as possible. Hold on, help is on the way.\\
                Thank you. We will try to keep the fire contained as much as we can.\\
            }
        \end{tabular}
    }
    \caption{Example Synthetic Chatter Generated by "Fire, Explosion" LoRA Adapter}
    \label{tab:example_fire}
\end{table}

\FloatBarrier
\section{Examples of the Logical Coherence Evaluation}
\label{sec:appendix_logical_coherence}

\begin{table}[htb]
    \centering
    \fbox{
        \begin{tabular}{c}
        \small
            \parbox{\textwidth}{
            \scriptsize
            \textbf{Context:}
        "vessel\_name": "ST PAUL",
            "vessel\_MMSI": "three six seven five one nine zero eight zero",
            "vessel\_call\_sign": "Whisky Delta Golf two seven six nine",
            "vessel\_type": "Motor Vessel",
            "vessel\_coordinate\_dms": "thirty-five degrees South, twenty degrees East",
            "compass\_direction": "south east",
            "closest\_place\_name": "Cape Agulhas",
            "distance\_to\_nearest\_place": "forty-seven",
            "closest\_place\_country": "South Africa",
            "distance\_to\_nearest\_port": "one hundred twenty-nine",
            "nearest\_port": "Cape Town harbour",
            "distance\_to\_nearest\_harbor": "one hundred twenty-nine",
            "nearest\_harbor": "Victoria \& Alfred Waterfront",
            "digit\_by\_digit": false,
            "can\_have\_cargo": null,
            "closest\_water\_body": null \\
            \textbf{Chatter:} \\
            "Mayday, Mayday, Mayday. This is fishing vessel RESURGENT, MMSI 454, call sign 454. We are a fishing vessel. We are in distress. We are listing. We need help. We are in danger of capsizing. Our location is seven degrees three three decimal four eight minutes South, one five nine degrees two five decimal six four minutes East. We are heading north east. Our closest place is Hakelake Rock, two two nautical miles. Our closest place country is Solomon Islands. We are in danger of capsizing. Our closest water body is Solomon Sea. We are in danger of capsizing. Our nearest harbor is Palunuhukura Harbour, four six nautical miles. We are in danger of capsizing. Our nearest port is Port Vila, one hundred fifty two nautical miles. We are in danger of capsizing. Our nearest port is Port Vila, one hundred fifty two nautical miles. We are in danger of capsizing. Our nearest port is Port Vila, one hundred fifty two nautical miles. We are in danger of capsizing. Our nearest port is Port Vila, one hundred fifty two nautical miles. We are in danger of capsizing. Our nearest port is Port Vila, one hundred fifty two nautical miles. We are in danger of capsizing. Our nearest port is Port Vila, one hundred fifty two nautical miles. We are in danger of capsizing. Our nearest port is Port Vila, one hundred fifty two nautical miles. We are in danger of capsizing. Our nearest port is Port Vila, one hundred fifty two nautical miles. We are in danger of capsizing. Our nearest port is Port Vila, one hundred fifty two nautical miles. We are in danger of capsizing. Our nearest port is Port Vila, one hundred fifty two nautical miles. We are in danger of capsizing. Our \\
            \textbf{Score:} 1 \\
            \textbf{Explanation:} Not a conversation between the Coast Guard and the vessel. Has a lot of repetitive statements.
            }
        \end{tabular}
    }
    \caption{Example Evaluation of a Synthetic Chatter with a\textit{Logical Coherence} Score of 1, Generated with the Vanilla LLM in Category "List-Danger of Capsizing"}
    \label{tab:example_logical_1}
\end{table}

\begin{table}[htb]
    \centering
    \fbox{
        \begin{tabular}{c}
        \footnotesize
            \parbox{\textwidth}{
                \textbf{Context:} \\
                "vessel\_name": "SEA PILOT", 
                "vessel\_MMSI": "three six eight one four four one two zero",
                "vessel\_call\_sign": "Whisky Delta Lima five seven five five",
                "vessel\_type": "Motor Vessel",
                "vessel\_coordinate\_dms": "sixty-three degrees North, one hundred sixty-one degrees West",
                "compass\_direction": "north west",
                "closest\_place\_name": "Black Point",
                "distance\_to\_nearest\_place": "four",
                "closest\_place\_country": "United States",
                "distance\_to\_nearest\_port": null,
                "nearest\_port": null,
                "distance\_to\_nearest\_harbor": null,
                "nearest\_harbor": null,
                "digit\_by\_digit": false,
                "can\_have\_cargo": null,
                "closest\_water\_body": "Black Cove"\\
            
                \textbf{Chatter:}
                    "Mayday, Mayday, Mayday. This is motor vessel SEA PILOT, MMSI number three six eight one four four one two zero, call sign Whisky Delta Lima five seven five five. We are located at the position sixty-three degrees North, one hundred sixty-one degrees West, four nautical miles north west of Black Point, Black Cove. We have a fire in the engine room, port side. We have three crew members on board and one of them is injured. We are experiencing difficulty controlling the fire and need assistance immediately. Over.", \\
                    "SEA PILOT, this is Coast Guard. We have dispatched a rescue boat to your location. ETA is thirty minutes. Can you provide more information about the extent of the fire and the condition of the injured crew member? Over.", \\
                    "The fire is spreading quickly. We have tried to use fire extinguishers, but it is not working. The injured crew member has smoke inhalation and is having difficulty breathing. Over.", \\
                    "Roger that. We will send a medical team along with the rescue boat. Can you tell us if you have enough life rafts and emergency beacons for all on board? Over.",
                    "Yes, we have enough life rafts and emergency beacons. Over.", \\
                    "Good. Stay calm and stay in the channel for further instructions. We will also send a fire fighting vessel to assist. Over.", \\
                    "Roger that. Thank you for your assistance. Over." \\

                    \textbf{Score:} 5
        }
        \end{tabular}
    }
    \caption{Example Evaluation of a Synthetic Chatter with a \textit{Logical Coherence} Score of 5, Generated with the LoRA Adapter "Fire, Explosion"}
    \label{tab:example_logical_5}
\end{table}

\FloatBarrier

\end{document}